\documentclass{article}
\usepackage{spconf,amsmath,graphicx}

\usepackage{overpic}
\usepackage{enumitem}
\usepackage{color}
\usepackage[breaklinks,colorlinks]{hyperref}
\usepackage{textpos}

\usepackage{pifont}
\usepackage{amsfonts}
\usepackage{caption}
\usepackage{subcaption}
\usepackage{epsfig}
\usepackage{booktabs}
\definecolor{turquoise}{cmyk}{0.65,0,0.1,0.3}
\definecolor{purple}{rgb}{0.65,0,0.65}
\definecolor{dark_green}{rgb}{0, 0.5, 0}
\definecolor{orange}{rgb}{0.8, 0.6, 0.2}
\definecolor{red}{rgb}{0.8, 0.2, 0.2}
\definecolor{darkred}{rgb}{0.6, 0.1, 0.05}
\definecolor{blueish}{rgb}{0.0, 0.3, .6}
\definecolor{light_gray}{rgb}{0.7, 0.7, .7}
\definecolor{pink}{rgb}{1, 0, 1}
\definecolor{greyblue}{rgb}{0.25, 0.25, 1}

\newcommand{\C}[1]{\small$\pm${#1}}

\newcommand{\Fig}[1]{Fig.~\ref{fig:#1}}

\newcommand{\Table}[1]{Table~\ref{tab:#1}}

\usepackage{blindtext}

\renewcommand{\paragraph}[1]{\vspace{1em}\noindent\textbf{#1}.}

\newcommand{\eg}{\textit{e.g.~}}
\newcommand{\ie}{\textit{i.e.~}}

\newcommand{\method}{\textsc{Face-TTS}\xspace}
\newcommand{\minus}{\text{-}}
\title{Imaginary Voice: Face-styled Diffusion Model for Text-to-Speech}

\name{Jiyoung Lee$^1$,
Joon Son Chung$^2$, 
Soo-Whan Chung$^3$}
\address{$^1$NAVER AI Lab, South Korea\\
         $^2$Korea Advanced Institute of Science and Technology, South Korea\\
         $^3$NAVER Cloud, South Korea}

\begin{document}
\ninept

\maketitle

\begin{abstract}

The goal of this work is zero-shot text-to-speech synthesis, with speaking styles and voices learnt from facial characteristics.
Inspired by the natural fact that people can imagine the voice of someone when they look at his or her face, we introduce a face-styled diffusion text-to-speech (TTS) model within a unified framework learnt from visible attributes, called \method. 
This is the first time that face images are used as a condition to train a TTS model.

We jointly train cross-model biometrics and TTS models to preserve speaker identity between face images and generated speech segments.
We also propose a speaker feature binding loss to enforce the similarity of the generated and the ground truth speech segments in speaker embedding space.
Since the biometric information is extracted directly from the face image, our method does not require extra fine-tuning steps to generate speech from unseen and unheard speakers.
We train and evaluate the model on the LRS3 dataset, an in-the-wild audio-visual corpus containing background noise and diverse speaking styles.
The project page is \url{https://facetts.github.io}.
\end{abstract}


\section{Introduction}
\label{sec:intro}

Text-to-speech~(TTS) is one of the core tasks in speech processing that generates speech waveform from a given text transcription.
Deep generative models have been introduced to produce high-quality spectral features from text sequences~\cite{oord2016wavenet,yamamoto2020parallel,wang2017tacotron}.
They have brought remarkable improvements in the quality of synthetic speech signals, compared to traditional parametric synthesis methods.

Recent works on diffusion models~\cite{sohl2015deep,ho2020denoising,dhariwal2021diffusion} have provided excellent generation results with outputs of high quality in various research fields such as image generation, video generation, and natural language processing.
For example, diffusion methods have achieved noteworthy results in image generation models; \eg DALLE-2~\cite{ramesh2022hierarchical}, Stable Diffusion~\cite{rombach2022high}.
Likewise, diffusion methods have shown impressive results in TTS compared to the previous generative methods, both in acoustic modeling~\cite{kim2022guided,kim2022guided2,popov2021grad} and in the vocoder~\cite{kong2020diffwave,koizumi2022specgrad}.

However, there are several unresolved challenges in the field of TTS.
One problem we address in this paper is expanding single speaker TTS model to multi-speaker TTS.
Since every person has different speaking styles, tones or accents, it is very challenging for the TTS model to learn various speaker styles.
The second and related problem is that a significant amount of target speakers' speech samples are required to generate voices of unseen speakers, even for multi-speaker TTS.
The variability of speaking styles means that the model must have access to significant amount of enrollment data to learn about each speaker. 
Since it is difficult to obtain clean enrollment utterances for each speaker, this raises the question ``what if face images can be used for enrollment instead of clean speech?''

\begin{figure}
    \centering
    \includegraphics[width=0.8\linewidth]{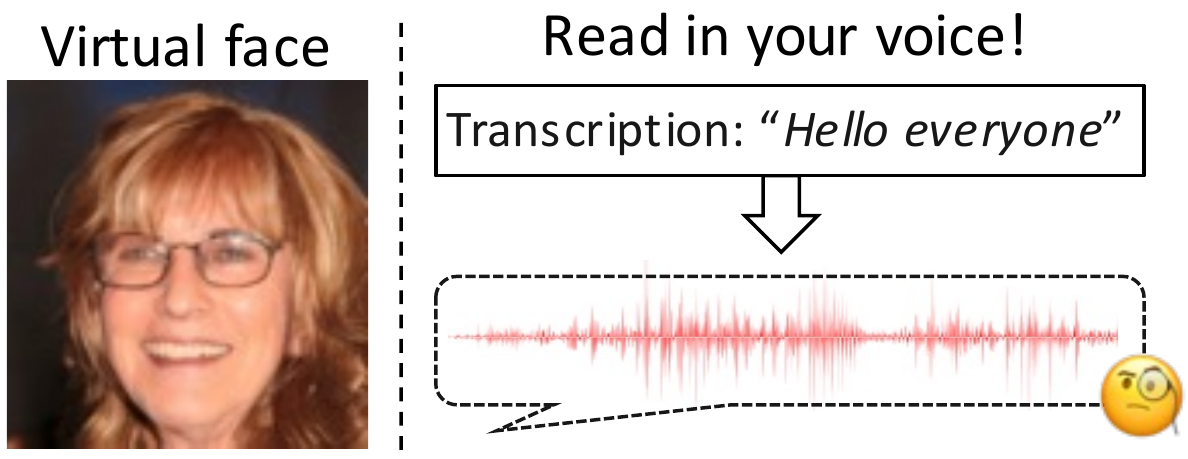}\hfill\\
    \vspace{-7pt}
    \caption{\method generates speech from a given text, conditioned on a face image. The face image is sampled from~\cite{rombach2022high}.}
    \label{fig:fig1}\vspace{-13pt}
\end{figure}

In~\cite{goto2020face2speech,wang2022residual}, the authors propose to leverage face images to control speaker characteristics of synthesised speech.
They train the face identity encoder to share a joint embedding space with the voice encoder, independently from the TTS model.
This approach enables generation of speech for unseen speakers without extra speaker adaptation.
However, these works do not use the face images as inputs when training the TTS models. 
Instead, the models are trained using speaker embeddings as the input, and the embeddings are swapped to face images only during inference.

\begin{figure*}[!t]
    \centering
    \includegraphics[width=\linewidth]{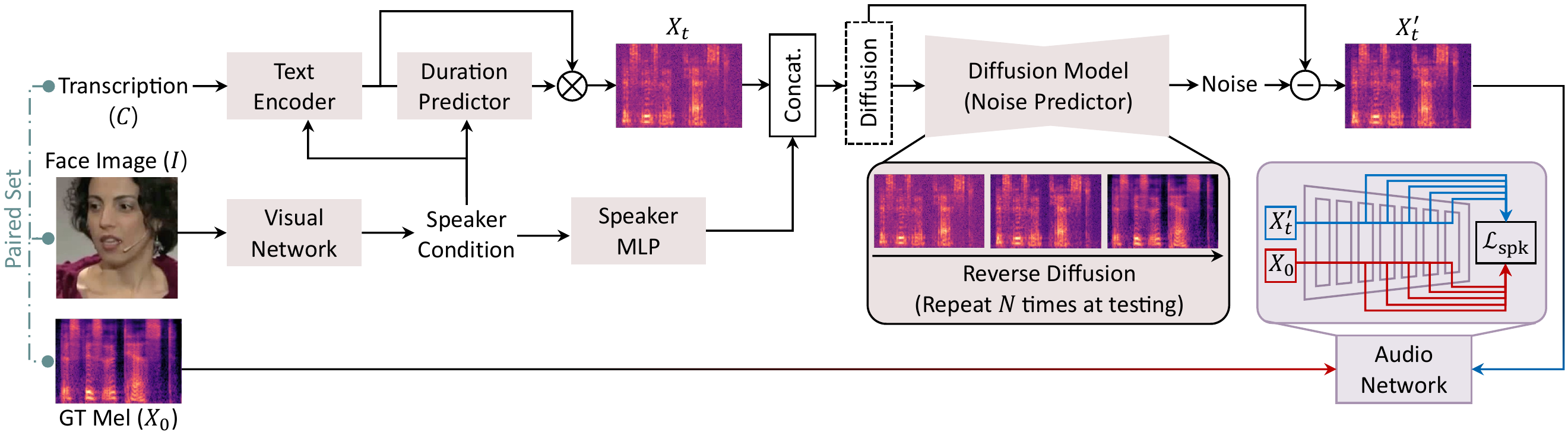}\\ 
    \vspace{-7pt}
    \caption{The overall configuration of \method. Given a text transcription and a face image, our method generates a speech sample using a diffusion model conditioned on face images to model speaker characteristics. The whole network except for audio network is trained end-to-end using the LRS3 dataset. Notice that the audio network are used only during training.}
    \label{fig:fig2}
    \vspace{-5pt}
\end{figure*}

In this paper, we propose a novel speech synthesis model, \method, which leverages face images to provide a robust characteristic of speakers.
In~\cite{nagrani2018learnable,chung2020seeing}, the authors have explored cross-modal biometrics and demonstrated that there is a strong correlation between voices and face appearances.
Inspired by this, we design a multi-speaker TTS model, where speaking styles are conditioned on face attributes.
While it is difficult to collect speech segments for the enrollment of every speaker, it is much easier to obtain face images. 
We enforce the matching of the identity of the face and the identity of the synthesised speech to train a robust cross-modal representation of speaking style. 
Our approach is capable of generating speech signals without speaker enrollment, which is advantageous in the zero-shot or few-shot TTS modeling.
Our backbone structure for the TTS model is derived from Grad-TTS~\cite{popov2021grad}, which learns acoustic features using the diffusion method.
Unlike other face-to-speech synthesis methods~\cite{goto2020face2speech,wang2022residual}, \method is trained end-to-end from the face encoder to the acoustic model, using in-the-wild datasets.
To the best of our knowledge, 
this is the first time that face images are used as a condition to train a TTS model.
We perform qualitative and quantitative tests to assess the speaker representations as well as the perceptual quality of the synthesised speech.
In addition, we verify through subject measures whether the synthesised speech fits well with the appearances of virtual humans who do not have their own voices as illustrated in~\Fig{fig1}.

\section{Related Work}
\vspace{-5pt}
\noindent\textbf{Text-to-speech.}
With the success of deep neural networks, the perceptual quality of synthesised speech is dramatically improved compared to the previous statistical parametric speech synthesis~\cite{zen2009statistical}.
In general, TTS models are composed of two modules; an acoustic model and a vocoder.
The acoustic model generates speech features (commonly mel-spectrogram) from text sequences, and the vocoder takes the features to generate speech waveform.
There have been many approaches using generative modeling methods~\cite{oord2016wavenet,yamamoto2020parallel,kong2020hifi}, and Tacotron-based models~\cite{wang2017tacotron,shen2018natural} incorporate a sequence-to-sequence model to transform the text sequences into the acoustic representations.
GAN-based models~\cite{yamamoto2020parallel,kong2020hifi} have brought innovative contributions to TTS in the last decade using adversarial training strategy.
Recently, another successful generative approach, diffusion-based methods~\cite{popov2021grad,kong2020diffwave,koizumi2022specgrad,kim2022guided}, have been proposed in speech synthesis, as the diffusion methods have proved their effectiveness in various generation tasks~\cite{nichol2021glide,rombach2022high,ramesh2022hierarchical}.
Compared to GAN-based models, diffusion methods have advantages in impressive results as well as distribution coverage, a fixed training objective, and scalability.

\noindent\textbf{Audio-visual biometrics.}
People instinctively co-relate others' facial appearances and their voices by learning through experiences, because face and voice provide related identity information~\cite{smith2016matching}.
In order to learn this correlation between faces and voices, several prior works~\cite{nagrani2018learnable,chung2020seeing} have tried to use self-supervised methods in the way that people learn from experience.
They have leveraged the fact that a face image and speech segment from a single-speaker video should have a common identity.
In~\cite{chung2020facefilter,gao2021visualvoice}, the authors have shown that visual identity has a strong correlation with the speaker identity by separating input signals using face images.
Various self-supervised losses have been considered to learn robust cross-modal embeddings for biometrics matching, such as cross-entropy loss~\cite{chung2020perfect}, contrastive loss~\cite{nagrani2018learnable} and disentanglement-based loss~\cite{nagrani2020disentangled}.
Motivated by these previous works, we leverage the cross-modal biometrics matching to foster conditions that reflect speaker-dependent characteristics for multi-speaker TTS model.

\vspace{-2pt}
\section{Face-TTS}
\label{sec:method}
\vspace{-3pt}

\subsection{Score-based Diffusion Model}\vspace{-2pt}
\method is based on a score-based diffusion model, specifically Grad-TTS~\cite{popov2021grad}, which consists of three main parts; (1)~text encoder, (2)~duration predictor, (3)~diffusion model. 
Formally, given a text transcription $C$ and a corresponding mel-spectrogram $X_0$ for training, the forward process progressively adds standard Gaussian noise to satisfy the following continuous stochastic differential equation (SDE)~\cite{song2020score}:
\vspace{-3pt}
\begin{equation}\label{eq:eq1}
    dX_t = -\frac{1}{2} X_t \beta_t dt + \sqrt{\beta_t} d W_t,
    \vspace{-3pt}
\end{equation}
where $W_t$ is the standard Brownian motion, and $\beta_t$ is a noise schedule.
In the reverse diffusion process, $X_0$ can be obtained from $X_t$ corresponding to the text as follows:
\vspace{-3pt}
\begin{equation}
    dX_t = - \Big(\frac{1}{2} X_t +\mathcal{S} (X_t,t) \Big)\beta_t dt + \sqrt{\beta_t} d \tilde{W}_t,
    \vspace{-3pt}
\end{equation}
where $\tilde{W}_t$ is the reverse-time Brownian motion and $\mathcal{S}(X_t,t)$ is a diffusion model that estimates the gradient of the log-density of noisy data $\nabla_{X_t} \log p_t(X_t)$.
Namely, we infer the speech $X_0$ from noisy data $X_t$ with $N$ steps by solving the SDE:
\vspace{-3pt}
\begin{equation}
    X_{t-\frac{1}{N}} = X_t + \frac{\beta_t}{N}\Big(\frac{1}{2} X_t +\mathcal{S} (X_t,t)\Big) + \sqrt{\beta_t} \tilde{W}_t,
    \vspace{-3pt}
\end{equation}
where $t \in \{\frac{1}{N}, \frac{2}{N}, \dots, 1\}$.
We note that $N$ is the number of steps to the discretised reverse process, and $t$ indexes a subsequence of time steps in the reverse process.
We follow most parts similar to the original methodology~\cite{popov2021grad} and explain the different points in the below sections. The overall architecture is illustrated in \Fig{fig2}.

\vspace{-3pt}
\subsection{Speaker Conditioning with Cross-modal Biometrics}
\vspace{-2pt}
In~\cite{popov2021grad,kim2020glow}, the authors do not utilise a speaker model for learning speaking styles in their TTS models, but prepare a pre-defined speaker codebook for each identity.
Thus, it is difficult to present a new speaker in their models, and it requires a challenging adaptation procedure to resolve this problem.
In~\cite{nachmani18fitting,chen2021adaspeech}, they prove that the speaker embedding precisely adjusts speaking styles in synthesised speech.
However, there still remains a problem.
Speaker embeddings usually represent excessive details of speakers, and it yields unstable training in the acoustic modeling of TTS.
Therefore, speaker embeddings should be generalised to represent speakers' voices in synthesised speech.

In this paper, we provide identity embedding from a face image as a conditioning feature on the TTS model for multi-speaker modelling.
Since the face embedding from the cross-modal biometric model represents the identity related to the voice, it is suitable to generate speech that matches face attributes.
Such face embedding does not contain a complex distribution of speakers, but only associative representations from voice and face, and it naturally generalises the speaker embedding and allows efficient multi-speaker modelling.
Given a mel-spectrogram $X$=$X_0$ and a face image $I$, the network is pre-trained to associate the same speaker identity from the different modalities, where the overall network consists of audio network $\mathcal{F}(X)$ and visual network $\mathcal{G}(I)$.

The visual network ingests a face image of the target speaker to produce a speaker representation.
Then the text encoder and the duration predictor estimate the statistics of acoustic features from given a text transcription and a face image.
In details, the text encoder generates acoustic features fit to text sequences, and the duration predictor colourises features with predicted speaking duration of the target speaker for the natural pronunciation.
During training, the diffusion process adds Gaussian noise on colourised features to make noisy data, and the diffusion model estimates the gradient of data distribution in noisy data to obtain the target audio.
Specifically, the speaker representation guides the diffusion model to estimate gradients optimal to generate synthesised speech in the speaker's voice.
We note that the network configuration follows~\cite{popov2021grad}.

However, to learn various speakers' characteristics for the multi-speaker TTS, the TTS model requires sufficient length of recorded speech for each person.
Previous works~\cite{kim2022guided,goto2020face2speech,wang2022residual} trained their models using audiobook dataset read by several speakers with enough lengths of utterances, where it is difficult to generalise models for unseen speakers.
To solve this problem, we suggest an effective strategy, a speaker feature binding loss, maintaining speaker characteristics of target voices in synthesised speech.
It allows \method to learn face-voice association from audio segments even with a short length.
Formally, latent embeddings from convolution layers of the audio network trained in cross-modal biometrics are extracted from synthesised speech and target voices, respectively.
The speaker feature binding loss $\mathcal{L}_{\text{spk}}$ train our \method model by minimising distances of two latent embedding sets as follows:
\vspace{-3pt}
\begin{equation}
    \mathcal{L}_{\text{spk}} = \sum\nolimits_B |\mathcal{F}_b(X_0)-\mathcal{F}_b(X'_{t})|,
    \vspace{-3pt}
\end{equation}
where $X_0$ is a mel-spectrogram from a target speaker's utterance and $X'_t$ is a denoised output from the network, and $B$ indicates the number of convolution blocks of audio network except for the first two convolution blocks.
We freeze the audio network not to be updated with this loss.
This training strategy enforces to form a speaker-related latent distribution of synthesised speech similar to that of the target speech.

\vspace{-3pt}
\subsection{Training \& Inference}
\vspace{-2pt}
In training session, \method learns multi-speaker speech synthesis through multiple training criteria.
To train text and duration encoders, we exploit the prior loss to estimate the mean from a normal distribution and the duration loss~\cite{kim2020glow} to control the duration of pronunciation using a monotonic alignment between speech and text sequences.
Diffusion loss trains the diffusion model to estimate the gradient of data distribution as in~\cite{popov2021grad}.
Our final training objective is described as:
\vspace{-3pt}
\begin{equation}
    \mathcal{L} = \mathcal{L}_{\text{prior}} + \mathcal{L}_{\text{duration}} + \mathcal{L}_{\text{diff}} +  \gamma\mathcal{L}_{\text{spk}},
    \vspace{-3pt}
\end{equation}
where $\gamma$ is empirically set to $1e \minus 2$.
We emphasise that the whole framework is trained end-to-end on LRS3 dataset obtained from in-the-wild environments.
Thanks to video in LRS3 with various angles and facial expressions, our \method is more robust to real-world face images than previous works~\cite{goto2020face2speech,wang2022residual} that only used the front view of a face image.

For inference, the trained \method samples a mel-spectrogram of utterance $X_0$ from the noisy data $X_t$ that is estimated by transcription with speaker condition by target speaker's face. 
The reverse diffusion process is repeatedly processed to estimate step-by-step noise gradually.
Finally, we used a pretrained vocoder to transform the estimated mel-spectrogram to a raw waveform.

\section{Experiments}
\vspace{-2pt}
\label{sec:experiments}
\subsection{Experimental Settings}
\vspace{-2pt}
\noindent\textbf{Datasets.}
LRS3~\cite{afouras2018deep} is an audio-visual dataset collated from TED videos, which has audio-visual pairs with corresponding text transcriptions.
We use the \textit{trainval} split for the training and the \textit{test} split for the evaluation, excluding speech samples shorter than 1.3 seconds.
Also, we pick out speech samples of speakers who have at least 10 seconds of audio in total. 
A total of 14,114 utterances and 2,007 speakers is used for training, 50 utterances for validation, and test set includes 412 speakers.
The widely used multi-speaker TTS dataset, such as LibriTTS~\cite{zen2019libritts}, has 550 seconds per speaker in average from well-recorded audio books, whereas LRS3~\cite{afouras2018deep} has a length of about 34 seconds extracted from real-world environments.
Therefore, it is extremely challenging to use LRS3 data to train TTS models.
We use the test split (448 samples) of LJSpeech~\cite{ljspeech17} to obtain text descriptions in the out-of-distribution for a fair comparison with previous works~\cite{popov2021grad, kim2020glow}.
The cross-modal biometric model~\cite{nagrani2018seeing} is re-implemented following to the same configuration of mel-spectrogram with vocoder~\cite{kong2020hifi}.
It is trained on VoxCeleb2~\cite{chung2018voxceleb2} dataset which contains 5,994 speakers in audio-visual pairs.

\noindent\textbf{Audio and image representation.}
The inputs to the network, including cross-modal biometric model, TTS model and vocoder, are the 128-dimensional mel-spectrogram extracted at every 10ms with 62.5ms frame length in 16kHz sampling rate.
For the image input, the face image is randomly sampled from each video and resized into 224$\times$224 pixels, same as in~\cite{chung2020seeing}.
The cross-modal biometric model (\ie audio and visual networks) embeds audio and face images onto 512-dimensional vectors.

\noindent\textbf{Evaluation protocols.}
In our experiments, the generated mel-spectrogram is synthesised into an audio waveform using HiFi-GAN as the vocoder.
We first report `Mel.+HiFi-GAN' to inform the degradation amount caused by the vocoder. 
In this case, mel-spectrogram of target speech is transformed into the waveform without synthesis process.
It is natural that it shows a little lower scores with the `Ground-Truth' result, and it can be the upper-bound score of synthesis results.
We perform mean opinion score (MOS) test, which is a common metric to measure subjective perceptual quality of synthesised speech.
A total of 17 participants are asked to judge the quality about the synthesis results in 5-scale: 1=Bad; 2=Poor; 3=Fair; 4=Good; 5=Excellent.
In the test, 10 utterances are randomly selected from the test set and synthesised using each model.
Additionally, we conduct two preference tests, 1) AB forced matching test; synthesised speech and two face images, 2) ABX preference test; two synthesised speech signals and one face image.
For the validity of our model for the virtual human speech generation, we perform the MOS test whether the synthesis outputs are harmonised with the face images generated from the recent image generation model~\cite{rombach2022high}.
Here, we provide choice options from 1 to 4, where the higher score means the synthesised speech is harmonised with the face image.
For the objective evaluation, we establish the 5-way cross-modal forced matching test through the cross-modal biometric model, which has to select the matching identity from synthesised speech and 5 face images.
In this matching test, we verify the synthesised speech represents similar identity appeared in a face image.

\noindent\textbf{Implementation details.}
For the fair comparison, we train Grad-TTS with LibriTTS and LRS3 datasets, respectively.
Also, our \method is trained using identity embeddings from audio inputs and face inputs, where both embeddings are obtained from cross-modal biometrics.
We follow the most of training configuration of Grad-TTS~\cite{popov2021grad}.
Since the visual network are initialised with pre-trained weights for the biometric matching task on VoxCeleb2, we give a smaller value ($1e\minus6$) as a initial learning rate for those networks. 
We notice that, except for audio and visual networks, the other networks are trained from scratch.
The computation time and flops are increased linearly, while more denoising step increases the audio quality.
Thus, we equally use 10 denoising or sampling steps to generate speech signals for the inference.

\begin{table}[t]
    \centering
    \begin{tabular}{lcc}
    \toprule
         \bf{Method} & \bf{Spk. ID} & \bf{5-scale MOS}  \\
        \midrule
        Ground Truth & - & 4.865\C{.001} \\
        Mel.+HiFi-GAN~\cite{kong2020hifi} (Upper bound) & - & 4.653\C{.035}\\
        \midrule
         Grad-TTS~\cite{popov2021grad}$\dagger$ (Seen) & Embed & 3.718\C{.318}\\
         \method (Seen) & Audio & 3.547\C{.331}\\
          \method (Seen) & Face & 3.706\C{.154}\\
          \method (Unseen) & Audio & 3.218\C{.249}\\
         \method (Unseen) & Face & 3.282\C{.219}\\
    \bottomrule
    \end{tabular}
    \vspace{-5pt}
    \caption{Subjective evaluation for comparison of audio quality with mean opinion score (MOS) metric. Grad-TTS$\dagger$ is trained on LibriTTS, and \method are trained on LRS3.}
    \label{tab:mos}\vspace{-5pt}
\end{table}
\begin{figure}[t]
    \centering
    \begin{subfigure}{\linewidth}
        \centering
        \includegraphics[width=0.82\linewidth]{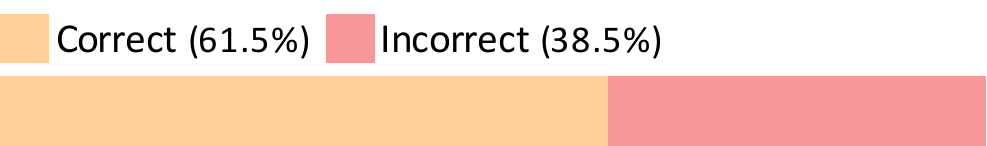}
    \caption{AB test}
    \end{subfigure}\hfill
    \begin{subfigure}{\linewidth}
        \centering
        \includegraphics[width=0.82\linewidth]{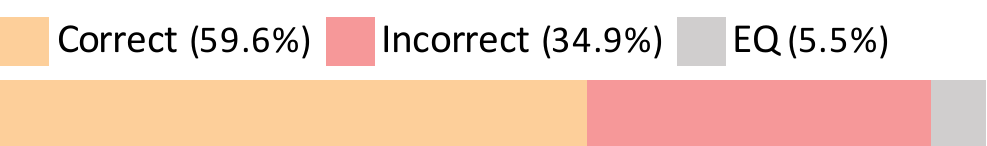}
    \caption{ABX test}
    \end{subfigure}\hfill
    \vspace{-9pt}
    \caption{Results of preference tests. (a) Preference for a face matching two synthesised utterances. (b) Preference for a synthesised utterance matching two face appearances.}\vspace{-9pt}
    \label{fig:fig3}
\end{figure}
\begin{table}[t]
    \centering
    \begin{tabular}{lcc}
    \toprule
         \bf{Method} & \bf{Spk. ID} & \bf{Acc. (\%)}  \\
        \midrule
        Mel.+HiFi-GAN~\cite{kong2020hifi} (Upper bound) &  - & 48.6\\
        \midrule
         Grad-TTS~\cite{popov2021grad} & Embed & 19.4 \\
         \method (w/o. $\mathcal{L}_{\text{spk}}$) & Face & 35.4  \\
         \method & Face & 38.0 \\
    \bottomrule
    \end{tabular}
    \vspace{-5pt}
    \caption{Speaker identification matching accuracy. Since Grad-TTS uses speaker id embedding, its model is evaluated with \textit{seen} speakers and our model is evaluated with \textit{unseen} speakers. Random accuracy is 20\%.}
    \vspace{-1pt}
    \label{tab:acc}
\end{table}
\begin{table}[t]
    \centering
    \begin{tabular}{lc}
    \toprule
        \bf{Test sample} & \bf{4-scale MOS}  \\
        \midrule
        LRS3 (Real) & 3.471\C{.291} \\
        Stable Diffusion~\cite{rombach2022high}+\method (Fake) & 2.941\C{.462} \\
    \bottomrule
    \end{tabular}
    \vspace{-5pt}
    \caption{Matching preference between virtual face images from Stable Diffusion~\cite{rombach2022high} and generated utterances with MOS.}\vspace{-5pt}
    \label{tab:virtual}
    \vspace{-8pt}
\end{table}

\subsection{Results}\vspace{-2pt}
\noindent\textbf{Audio quality.}
We brought pre-trained parameters of Grad-TTS from the author for the comparison, and it had been trained in the LibriTTS dataset for multi-speaker TTS.
In our preliminary experiment, we empirically found that Grad-TTS trained on LRS3 dataset showed competitive perceptual quality.
Therefore, we evaluated the Grad-TTS trained on LibriTTS as a comparison following authors' official implementation, and we re-sampled generated audio from Grad-TTS from 22.05kHz to 16kHz.
\method with audio speaker ID was fully trained with the audio network in cross-modal biometrics instead of the visual network.
In~\Table{mos}, the result indicates that \method using face images shows competitive audio quality to Grad-TTS trained on clean speech dataset under the seen speaker condition.
We observed that our \method can generate audio of fine quality (\ie above 3 score) even for unseen speakers.
Furthermore, there is a little difference in the performance between the models using face or audio as conditions.
Compared audio-conditioned models, face conditioning has brought more fine-grained audio quality, because the face represents robust identity compared to the speech influenced by recording environments.

\noindent\textbf{Speaker verification.}
We further evaluate the speaker verification task with generated utterances and face images. 
First, AB and ABX preference tests are performed on human evaluators.
To evaluate under more challenging conditions, we conducted the experiment with gender unified. 
That is, the face or audio in the two cases to be selected were selected from samples of the same gender.
The evaluators selected a correct answer rate of about 60\% as reported in \Fig{fig3}. 
Furthermore, \Table{acc} shows 5-way cross-modal speaker matching accuracy for objective evaluations on the LRS3 dataset.
Following their official implementation, we train Grad-TTS~\cite{popov2021grad} on the LRS3 dataset for this experiment.
Although the Grad-TTS trained on the LRS3 shows competitive audio quality with ours, capturing the speakers' characteristics in the sound with the settled speaker embedding seems challenging in the Grad-TTS.
Moreover, our speaker loss improves the matching performance 2.6\% than \method without the loss, training the diffusion model to sample the utterance, which is more proper to the target face.
However, it still has room to improve the performance up to the result in the first row (Mel.+HiFi-GAN). We remain it as future work.

\noindent\textbf{Virtual speech generation.}
To demonstrate the utility of our \method, we synthesised speech with virtual face images generated from \cite{rombach2022high}.
\Table{virtual} reports the subjective evaluation of 4 points Likert-scale measurement: 1=Bad; 2=Neutral; 3=Good; 4=Excellent.
We had assessors evaluate virtual faces without knowing they were mixed.
As the baseline, we also evaluate the preference of ground-truth face-voice pairs, which are randomly selected on the LRS3 dataset.
Surprisingly, people gave `Good' score on average, in that utterance from our \method is well matched with virtual face images.

\vspace{-3pt}
\section{Conclusion}
\label{sec:conclusion}
\vspace{-2pt}
In this work, we proposed \method for multi-speaker text-to-speech synthesis with speaker identity conditioned by a face image.
For this goal, we leveraged the cross-modal biometric to specify the speaker characteristics from the face for the diffusion-based TTS model, instead of enrolled speech.
To jointly train the two modules for enhancing generation performance, we introduced the speaker feature binding loss to maintain speaker consistency between synthesised speech and reference speech.
Both quantitative and qualitative evaluations confirmed the high-quality generation of \method, particularly containing good representations of target speakers' voices.
Moreover, we demonstrated that \method is effective for using realistic-sounding voices of virtual humans, which introduces an interesting application to the emerging field.

\vspace{2pt}
\noindent\textbf{Acknowledgments.}
The NAVER Smart Machine Learning (NSML) platform~\cite{NSML} has been used in the experiments.

\clearpage
\bibliographystyle{IEEEbib}
\bibliography{refs}

\end{document}